\documentclass{article}

\PassOptionsToPackage{numbers}{natbib}

\usepackage[preprint]{neurips_2023}

\usepackage[utf8]{inputenc} 
\usepackage[T1]{fontenc}    
\usepackage{url}            
\usepackage{booktabs}       
\usepackage{amsfonts}       
\usepackage{nicefrac}       
\usepackage{microtype}      
\usepackage{xcolor}         
\usepackage[colorlinks=true,
            linkcolor=red,
            citecolor=blue,
            filecolor=blue,
            urlcolor=blue]{hyperref}
\usepackage{enumitem}
\setlist[itemize]{leftmargin=*}

\usepackage{ntheorem}

\usepackage{amsmath}
\allowdisplaybreaks[4]
\usepackage{amssymb}
\usepackage{wrapfig}
\usepackage{multirow}
\usepackage[ruled,lined]{algorithm2e}
\usepackage{subfigure}
\usepackage{graphicx}
\usepackage{float}
\usepackage{arydshln}

\usepackage{colortbl}
\usepackage{xcolor}

\usepackage{footmisc,graphicx}
\newcommand*\wine{\protect\includegraphics[scale=0.15]{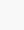}}
\DefineFNsymbols*{woexl}{\wine\dagger\ddagger\S\P\|%
{**}{\dagger\dagger}{\ddagger\ddagger}}
\setfnsymbol{woexl}

\title{Building Accurate Translation-Tailored LLMs with Language Aware Instruction Tuning}

\author{%
  Changtong Zan\\
  China University of Petroleum (East China)\\
  \texttt{zanct@s.upc.edu.cn} \\
  \And
  Liang Ding\includegraphics[scale=0.15]{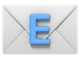}\thanks{\includegraphics[scale=0.15]{figure/mail.png} Liang Ding and Weifeng Liu are the corresponding authors.}\\
  The University of Sydney\\
  \texttt{liangding.liam@gmail.com} \\
  \And
  Li Shen \\
  JD Explore Academy \\
  \texttt{mathshenli@gmail.com}
  \And
  Yibing Zhan \\
  JD Explore Academy \\
  \texttt{zhanyibing@jd.com}
  \And
  Weifeng Liu\includegraphics[scale=0.15]{figure/mail.png} \\
  China University of Petroleum (East China)\\
  \texttt{liuwf@upc.edu.cn}
  \And
  Dacheng Tao \\
  Nanyang Technological University \\
  \texttt{dacheng.tao@ntu.edu.sg}
}

\begin{document}

\maketitle

\begin{abstract}
\label{abstract}
Translation-tailored Large language models (LLMs) exhibit remarkable translation capabilities, even competing with supervised-trained commercial translation systems. However, \textbf{off-target translation} remains an unsolved problem, especially for low-resource languages, hindering us from developing accurate LLMs-based translation models.
To mitigate the off-target translation problem and enhance the performance of LLMs on translation, recent works have either designed advanced prompting strategies to highlight the functionality of translation instructions or exploited the in-context learning ability of LLMs by feeding few-shot demonstrations. 
However, these methods essentially do not improve LLM's ability to follow translation instructions, especially the language direction information.
In this work, we design a two-stage fine-tuning algorithm to improve the instruction-following ability (especially the translation direction) of LLMs.
Specifically, we first tune LLMs with the maximum likelihood estimation loss on the translation dataset to elicit the basic translation capabilities. In the second stage, we construct instruction-conflicting samples by randomly replacing the translation directions with a wrong one within the instruction, and then introduce an extra unlikelihood loss to learn those samples.
Experiments on IWSLT and WMT benchmarks upon the LLaMA model spanning 16 zero-shot directions show that, compared to the competitive baseline --  translation-finetuned LLama, our method could effectively reduce the off-target translation ratio (averagely -53.3\%), thus improving translation quality with average +5.7 SacreBLEU and +16.4 BLEURT. Analysis shows that our method could preserve the model's general task performance on AlpacaEval.
Code and models will be released at \url{https://github.com/alphadl/LanguageAware_Tuning}.
\end{abstract}
\section{Introduction}
\label{introduction}
Large language models (LLMs) have demonstrated excellent performance on a wide range of NLP tasks, including reasoning~\cite{wei2022cot}, summarization~\cite{wang-2023-gpt4summarization}, translation~\cite{hendy2023gpt4mt}, understanding~\cite{zhong2023chat}, and evaluation~\cite{Lu2023EAPrompt}, etc. 
LLMs exemplified by GPT-3~\cite{brown2020gpt3}, OPT~\cite{zhang2022opt}, LLaMA~\cite{touvron2023llama}, and LLAMA2~\cite{touvron2023llama2}, leverage large-scale monolingual data through pertaining with the causal language modelling task and exhibit strong zero-shot capabilities with few demonstration examples. 
Instruction tuning~\cite{naturalinstructions, wei2021flan} further elicits the capacity of LLMs to address general tasks directly with proper guidance, such as task definition. 
Nevertheless, due to the huge cost to call the state-of-the-art LLM, like GPT-4~\cite{openai2023gpt4}, it is attractive to explore strategies for effectively fitting suitably-sized LLMs into specific tasks~\cite{xu2023paradigm_shift, pmlr-v202-fu23d}.

\begin{wrapfigure}{hr}{0.5\textwidth} 
    \flushleft
    \centering
\includegraphics[width=0.47\textwidth]{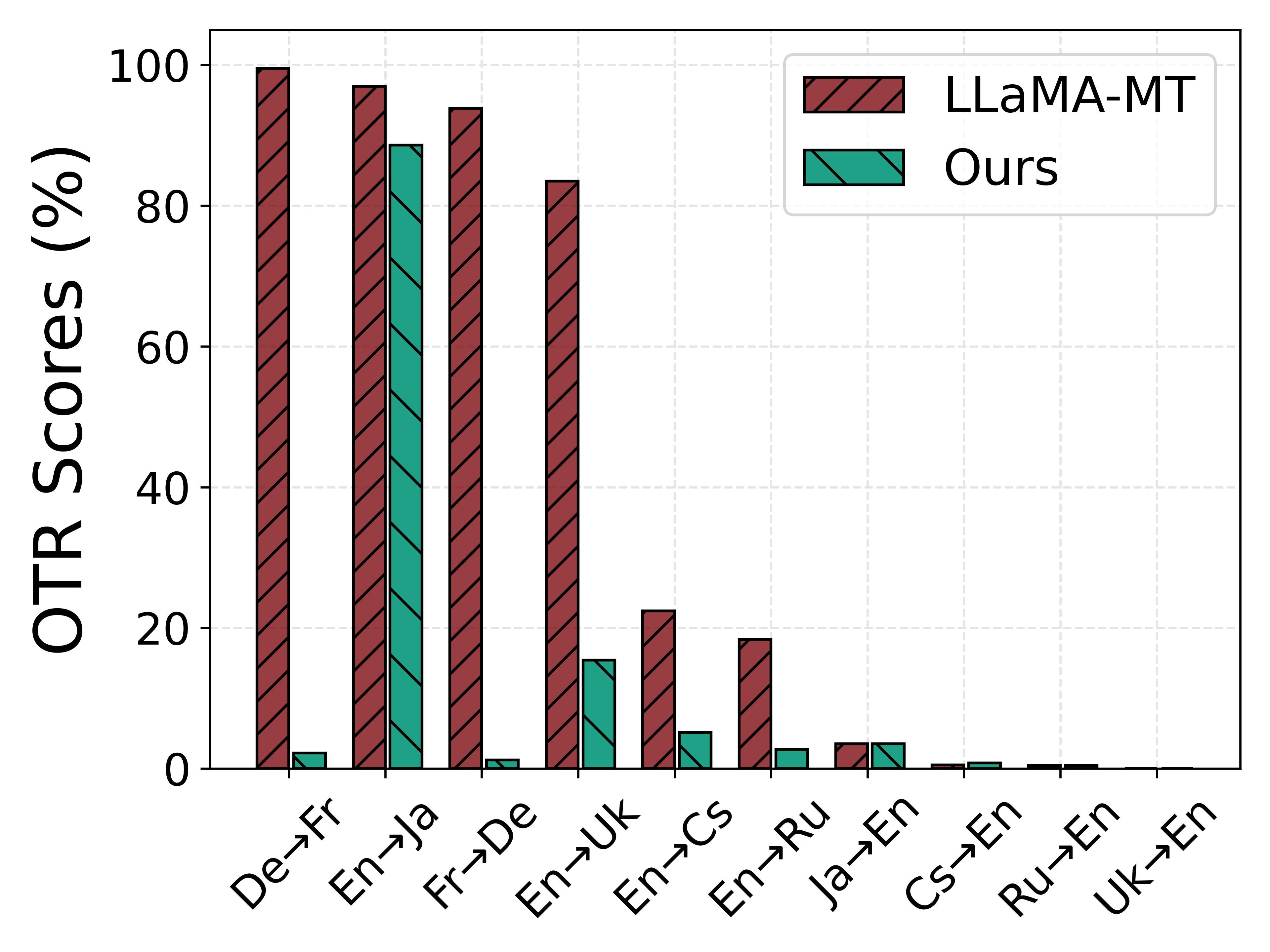}
    \caption{
    \textbf{Off-target translation ratio (OTR \% $\downarrow$) in zero-shot translation of WMT dataset}. We present the comparison between LLaMA-MT, a LLaMA fine-tuned on the translation dataset, and our model.
    }
    \label{fig:intro_example}
\end{wrapfigure}

In the field of zero-shot translation (ZST)~\cite{gu-etal-2019-improved,chen-2023-target,zan2023unlikelihood}, the goal of this task is to translate sentences from a source language to a target language, where 1) the direct mappings between source and target languages lack in the training data, or 2) the target or source language themselves have not appeared during training.
Addressing the ZST problem is both vital and challenging, especially for paired-data-hungry low-resource languages.
Recent research demonstrates that LLMs tuned on translation data can achieve good translation performance by configuring a suitable task instruction~\cite{zeng2023tim, liu2023instruction_position, xu2023paradigm_shift}. 
However, as illustrated in Figure~\ref{fig:intro_example}, our preliminary study shows that, when tackling zero-shot directions, LLM heavily encounters the off-target problem, for example,
in De$\rightarrow$Fr, the off-target ratio reaches 99.5\%\footnote{The effectiveness of our method can be significantly shown consistently besides the En$\rightarrow$Ja direction, the reason for the relatively weak improvement in Japanese may be due to Llama's vocabulary overly compressing non-Western languages.}.

We attribute this problem to the reason that training LLMs with the fashion of predicting the next token may lead to overlooking the information contained in instructions.  
Previous studies~\cite{Peng2023ChatGPT4MT,xu2023paradigm_shift} indicate that introducing more informative prompts during inference, such as preemptively translating prompts into the target language or incorporating few-shot demonstrated samples, can be beneficial. 
\citet{sennrich2023cd_ot} modify the decoding by introducing language-contrastive samples to constrain the decoding process, thus alleviating the off-target problem.
Different from the above approaches that focus on maximizing the utilization of LLMs for translation, our motivation is to fundamentally improve the instruction-following ability (especially the language-aware translation direction) of LLMs themselves.

In this paper, we introduce a simple-and-effective two-stage fine-tuning algorithm to enhance the effect of instruction in translation-tailored LLMs. This is accomplished by introducing unlikelihood loss on instruction-conflicting samples in which the translation sequence pairs deviate from the prescribed tasks associated with the given instructions.
In the first stage, we fine-tune the LLMs using a multilingual translation dataset. This pre-tuning process serves the purpose of unlocking the translation capabilities inherent in LLMs.
In the second stage, we build upon the pre-tuned model by incorporating translation data along with instruction-conflicting samples. We create instruction-conflicting samples by randomly replacing the translation directions with a wrong one.  
These data are used to further train the model, leveraging the unlikelihood training paradigm.
Our approach can be viewed as emphasizing the effect of instructions, thereby guiding the model to produce translation in the correct language.
In the experiments, we apply our method to fine-tune the LLaMA model. The results reveal substantial reductions in the off-target translation ratio, with improvements of -92.2\% and -29.9\% on the IWSLT and WMT benchmarks, respectively. 
This leads to notable enhancements in translation quality, as evidenced by increases of average +23.0/ +12.4 BLEURT and +5.2/ +6.1 SacreBLEU in IWSLT/ WMT datasets.
Also, our method maintains the translation capability on supervised directions.
The main contributions are as follows: 
\begin{itemize} 
    \item We reveal the heavy off-target problem in LLM-based zero-shot translation settings, and we attribute this problem to the weak instruction (translation direction) following ability.
    \item To fundamentally improve the translation direction following ability, we introduce a two-stage fine-tuning algorithm for LLMs that leverages instruction-conflicting samples. 
    \item Extensive experiments illustrate the effectiveness of our approach in mitigating the off-target problem and producing better translations. Analyses show that our method will not affect the general ability of LLM, e.g., general task performance on AlpacaEval.
\end{itemize} 
\section{Preliminary}
\label{preliminary}
\paragraph{Instruction Tuning} Instruction tuning aims to refine LLMs through fine-tuning a diverse collection of data characterized by explicit instructions. This refinement process significantly enhances the zero-shot performance on previously unseen tasks~\cite{wei2021flan}. 
Each instance in the instruction tuning dataset comprises three fundamental components: 
1) \textit{Instruction}: This is a textual representation that describes NLP tasks in natural language.
2) \textit{Input} (optional): Supplementary contextual information that provides additional context for the given task. 
3) \textit{Output}: The expected response that LLMs should generate. 
During the tuning process, the model is trained using a teacher-forcing approach~\cite{cho2014teacher_forcing}. It models the distribution of output tokens conditioned on the instruction and, optionally, the input. This training methodology empowers the model to understand and follow instructions effectively. 
Subsequently, the instruction-tuned model is capable of directly performing unseen tasks by following the appropriate task instructions in a zero-shot manner. 
In this study, our primary focus is translation-tailored LLMs, where we fine-tune LLMs on paired multilingual translation data. 

\paragraph{Unlikelihood Training}~\citet{Welleck2020UL} explores a novel approach that encourages the model to assign lower probabilities to improbable generations, in contrast to the traditional likelihood training, which focuses on the overall probability distribution of correct sequences.
The general training framework comprises two types of updates:
1) Likelihood updates on ground-truth sequences, ensuring they are assigned high probabilities.
2) Unlikelihood updates on negative candidate sequences, preventing them from receiving excessively high probabilities. 
We extend this approach to the domain of zero-shot translation based on translation-tailored LLMs. We introduce instruction-conflicting samples for unlikelihood updates, thereby emphasizing the impact of translation instructions (especially the translation direction and language) and addressing off-target problems.
\begin{figure*}[tb] 
    \centering
    \includegraphics[width=0.96\textwidth]{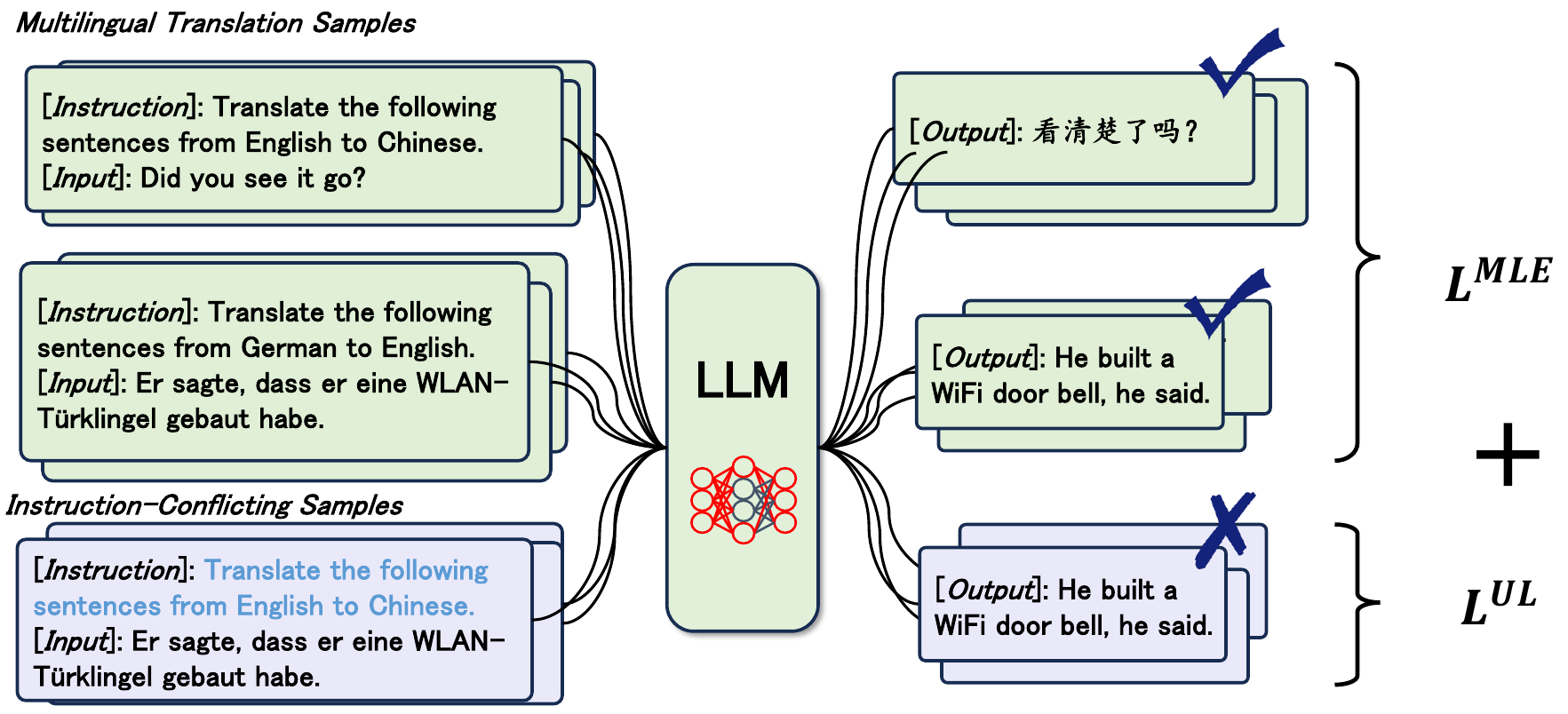}
    \caption{
    \textbf{Overview of our fine-tuning framework for zero-shot translation}. (a) In the first stage, we perform pre-tuning on LLMs using MLE loss on multilingual translation samples, focusing on unlocking the translation ability of LLMs. (b) Subsequently, we introduce instruction-conflicting samples by randomly substituting the instruction component with a different one. We then train the model with MLE loss $\mathcal{L}^{MLE}$ on translation data and incorporate an unlikelihood loss $\mathcal{L}^{UL}$ on the instruction-conflicting samples.
    }
    \label{fig:main_figure}
\end{figure*} 
\section{Methodology}
\label{methodology}

\subsection{Pre-Tuning on Multilingual Translation Samples}
To unlock the translation capabilities of LLM, we use the multilingual translation examples for the first stage pre-tuning. 
Formally, an LLM is pre-tuned with a collection of instruction samples $\mathcal{D}=\left\{\mathcal{D}_1, ..., \mathcal{D}_i, ..., \mathcal{D}_N \right\}$ covers $N$ language pairs. Here, $\mathcal{D}_i$ denotes a translation parallel corpus of the $i$-th language pair. 
As depicted in Figure~\ref{fig:main_figure}, in training stage 1, the model is trained to predict output based on provided instructions, such as ``Translate the following sentences from English to Chinese.'' and ``Translate the following sentences from German to English.'', and corresponding input like ``Did you see it go?'' and ``Er sagte, dass er eine WLAN-Türklingel gebaut habe.''. The likelihood training objective is applied:
\[
\begin{split}
\label{eq:likelihood_loss} 
\mathcal{L}^{\mathrm{MLE}}_\mathcal{D}(\theta)= -\sum_{\mathcal{D}_i \in \mathcal{D}}\sum_{(\boldsymbol{ins}, \boldsymbol{x}, \boldsymbol{y}) \in \mathcal{D}_i} \log{P(\boldsymbol{y} | \boldsymbol{ins}, \boldsymbol{x}; \theta)}, \nonumber 
\end{split}
\]
where $(\boldsymbol{ins}, \boldsymbol{x}, \boldsymbol{y})$ denotes task instruction, input, and output respectively. In the context of translation samples, $\boldsymbol{x}$ is the source sentence, and $\boldsymbol{y}$ is the target sentence. 
$\theta$ represents the trainable model parameters. 
Consequently, the model gets some capability to execute translation tasks by adhering to provided instructions. 

\subsection{Unlikelihood Training with Instruction-Conflicting Samples}
\label{sec:ul}
We enhance the zero-shot translation ability through a dual optimization approach, incorporating Maximum Likelihood Estimation (MLE) loss on multilingual translation samples and unlikelihood loss on instruction-conflicting samples.

\paragraph{Instruction-Conflicting Samples} 
To mitigate the off-target problem with unlikelihood training, we build the negative candidate samples by replacing the instruction with another different one while keeping the input and output not changing. 
We call this type of samples \textit{instructiong-conflicting samples} as the translation pairs deviate from the prescribed tasks associated with the given instructions.
As shown in Figure~\ref{fig:main_figure}, we sample a sample $(\boldsymbol{ins}, \boldsymbol{x}, \boldsymbol{y})$ from the instruction dataset $\mathcal{D}$, where $\boldsymbol{ins}$ is ``Translate the following sentences from German to English.'', $\boldsymbol{x}$ is `` Er sagte, dass er eine WLAN-Türklingel gebaut ha'', and $\boldsymbol{y}$ is ``He built a WiFi door bell, he sai''. 
Then, we randomly select another sample of a different task $(\widetilde{\boldsymbol{ins}}$, e.g. ``Translate the following sentences from English to Chinese'', and replace the original correct $\boldsymbol{ins}$ to get the instruction-conflicting sample, e.g. ``\textbf{\textit{[Instruction]}}: Translate the following sentences from English to Chinese. \textbf{\textit{[Input]:}} Er sagte, dass er eine WLAN-Türklingel gebaut ha'' in example.

\paragraph{Unlikelihood Training with Instruction-Conflicting Samples} 
Based on the instruction-conflicting samples, we generalize the unlikelihood training to zero-shot translation of translation-tailored LLMs. 
We feed instruction samples into the model trained after stage 1, optimizing the unlikelihood loss:
\[
\begin{split}
\label{eq:unlikelihood_loss} 
\mathcal{L}^{\mathrm{UL}}_\mathcal{D}(\theta)= -\sum_{\mathcal{D}_i \in \mathcal{D}}\sum_{(\boldsymbol{ins}, \boldsymbol{x}, \boldsymbol{y}) \in \mathcal{D}_i} \log{1-P(\boldsymbol{y} | \widetilde{\boldsymbol{ins}}, \boldsymbol{x}; \theta)},  \nonumber 
\end{split}
\]
where $\widetilde{\boldsymbol{ins}}$ is one of corresponding negative instructions for $\boldsymbol{ins}$. Thus, the overall objective function in unlikelihood training consists of mixing the likelihood and unlikelihood loss:
\begin{eqnarray}
\mathcal{L}_\mathcal{D}(\theta) = \mathcal{L}^{\mathrm{MLE}}_\mathcal{D}(\theta) + \alpha \mathcal{L}^{\mathrm{UL}}_\mathcal{D}(\theta),   \nonumber 
\end{eqnarray}
where $\alpha$ is the mixing hyper-parameter. 
\section{Experiments}
In this section, we conduct a series of experiments spanning 16 zero-shot translation directions to assess the effectiveness of our algorithm. 
\subsection{Experimental Setup}
\label{sec:experimental_setup}
\paragraph{Datasets}
We consider the following two widely-used datasets:
\begin{itemize}
 \item \textbf{WMT}: Following~\citet{jiao2023parrot, liu2023instruction_position}, we use the development sets from WMT2017 to WMT2020 for instruction tuning, including four language directions: En$\leftrightarrow$Zh and En$\leftrightarrow$De. The WMT dataset encompasses 51k translation sentence pairs.
 Then, we evaluate translation performance on WMT22 test sets, including En$\leftrightarrow$Cs, En$\leftrightarrow$Ja, En$\leftrightarrow$Ru, En$\leftrightarrow$Uk, Fr$\leftrightarrow$De. All these translation language pairs do not exist in the training set, thus allowing for the evaluation of zero-shot translation performance.  
 \item \textbf{IWSLT}: We collect the IWSLT dataset and focus on translation performance between non-English languages. 
 For fine-tuning using multilingual translation data, we randomly select 12k sentence pairs from the train set of IWSLT 2017~\cite{iwslt-2017-overview}, spanning six directions: En$\leftrightarrow$De, En$\leftrightarrow$Zh, En$\leftrightarrow$Ko. 
 We utilize Flores-200~\cite{nllb2022} devtest sets for evaluation on zero-shot translation directions, including Zh$\leftrightarrow$De, Zh$\leftrightarrow$Ko, De$\leftrightarrow$Ko. The Flores-200 comprises 1012 sentences from English Wikipedia covering multi-domain and then translated into about 200 languages by professional translators. 
\end{itemize}

\paragraph{Baselines} 
We leverage 7B size LLaMA as the backbone and consider the following baselines:
\begin{itemize}
     \item \textbf{LLaMA~\cite{touvron2023llama}}: LLaMA serves as the foundation model, having undergone training on a corpus of trillion tokens. We employ pretrained 7B size LLaMA directly for inference.
     \item \textbf{LLaMA-MT}: We fine-tune the LLaMA solely on multilingual translation samples, following the same procedure as the pre-tuning stage in our algorithm. Following \citet{jiao2023parrot}, we format translation sentence pairs into unified translation instructions.
     \item \textbf{Post-Ins~\cite{liu2023instruction_position}}: Following \citet{liu2023instruction_position}, we switch the positions of instruction and input of prompt, where the model pays more attention to the instruction. 
     \item \textbf{Prompt in the target language (PTL)}: 
     Instead of using the English prompt during inference, we translate the prompt into the target language during inference, which could provide more guidance information. Our inference leverages LLaMA-MT.
     \item \textbf{$K$-shot}: In-context learning~\cite{brown2020gpt3} has proven to be an effective way to prompt LLMs performance. We report the few-shot performance for comprehensive comparison, including 1-shot and 5-shot. LLaMA-MT is used for inference.
     \item $\mathbf{\mathcal{C}_{lang}}$~\cite{sennrich2023cd_ot}: Following \citet{sennrich2023cd_ot}, we propose to decode by contrasting the translation sentence with language-contrastive input and $\lambda_{lang}$ 0.5. Our inference relies on LLaMA-MT, employing a greedy decoding strategy.
\end{itemize}

\paragraph{Model Training} 
We conduct experiments on the {\tt Huggingface Transformers}~\cite{wolf-etal-2020-transformers} toolkit. All models are trained on Tesla-A100 GPUs. 
During the pre-tuning phase, we set the learning rate (lr) to be \texttt{2e-5}, the warmup ratio as \texttt{0.03}, and the batch size at \texttt{128}. For the IWSLT dataset, we performed training over \texttt{3} epochs, while for the WMT dataset, training was conducted for \texttt{1} epoch.
During the second stage of training, we set the mixing parameter denoted as $\alpha$ to \texttt{0.05}, the lr to \texttt{2e-6}, the batch size to \texttt{8}, and the training step to \texttt{100}. 
We use the final model for evaluation.

\paragraph{Evaluation}
We adopt \textbf{SacreBLEU}~\cite{post-2018-sacrebleu} to evaluate the translation accuracy, where translations are generated with a beam size of 4, a temperature of 0.1, and a top\_p of 0.9. 
Besides, we compute the ratio of wrong language translation in the generated outputs, i.e. off-target translation ratio (\textbf{OTR}), with publicly available language detector\footnote{\url{https://fasttext.cc/docs/en/language-identification.html}}~\cite{joulin2016fasttext,joulin2016bag}. 
Following~\citet{icml-v202-unreasonable-fst-translation}, we also use \textbf{BLEURT} \cite{sellam2020bleurt} to assess the translation quality with BLEURT-20 checkpoint\footnote{\url{https://github.com/google-research/bleurt}}.

\begin{table*}[t]
\centering
\label{tab:wmt}
\renewcommand{\arraystretch}{1.15} 
\resizebox{\linewidth}{!}{
\begin{tabular}{lccccccccccc}
\toprule 
\multirow{2}{*}{\textbf{Models}} & \multicolumn{2}{c}{\textbf{Cs-En}} & \multicolumn{2}{c}{\textbf{Ja-En}} & \multicolumn{2}{c}{\textbf{Ru-En}} & \multicolumn{2}{c}{\textbf{Uk-En}} & \multicolumn{2}{c}{\textbf{Fr-De}} & \multirow{2}{*}{\textbf{AVG}} \\ \cmidrule(lr){2-3} \cmidrule(lr){4-5} \cmidrule(lr){6-7} \cmidrule(lr){8-9} \cmidrule(lr){10-11}
& \bf $\leftarrow$     & \bf $\rightarrow$    & \bf $\leftarrow$     & \bf $\rightarrow$    & \bf $\leftarrow$     & \bf $\rightarrow$    & \bf $\leftarrow$     & \bf $\rightarrow$    & \bf $\leftarrow$     & \bf $\rightarrow$    & \  \\  \hline \addlinespace[1pt] 
& \multicolumn{10}{c}{\textbf{\textit{SacreBLEU Score $\uparrow$}}}     &  \\
\textbf{LLaMA}     & ~~1.0  & ~~1.9  & ~~0.3  & ~~2.4  & ~~0.8  & ~~2.5  & ~~0.8  & ~~3.4  & ~~1.7  & ~~1.4  & ~~\underline{1.6}  \\
\textbf{LLaMA-MT}  & 14.7 & 35.5 & ~~1.1  & 10.9 & 16.5 & 33.2 & ~~5.7  & 31.4 & ~~4.4  & ~~3.3  & \underline{15.7} \\
\textbf{Post-Ins}  & \textbf{17.3} & \textbf{37.5} & ~~\textbf{1.4}  & \textbf{11.9} & \textbf{19.5} & \textbf{34.9} & ~~8.0  & \textbf{33.1} & 11.0 & ~~3.2  & \underline{17.8} \\
\textbf{PTL}       & 13.1 & 35.5 & ~~1.0 & 10.9 & 13.3 & 33.2 & \textbf{14.1} & 31.4 & ~~8.5 & 10.6 & \underline{17.2} \\
\textbf{1-shot}    & 14.3 & 33.6 & ~~1.0 & ~~9.7 & 15.9 & 31.4 & ~~9.9  & 29.3 & ~~4.0 & ~~3.3 & \underline{15.2} \\
\textbf{5-shot}    & 15.3 & 32.6 & ~~1.1 & ~~8.7 & 16.3 & 30.1 & 11.8 & 28.6 & ~~4.0 & ~~3.2 & \underline{15.2} \\
$\mathcal{C}_{lang}$  & ~~2.3 & 31.4 & ~~0.2 & ~~8.7 & ~~2.1 & 29.9 & ~~1.4 & 27.8 & ~~1.2 & ~~0.6 & \underline{10.5} \\ 
\rowcolor{green!10} \textbf{Ours}      & 17.1 & 36.4 & ~~\textbf{1.4}  & 10.6 & 18.4 & 34.8 & \textbf{14.1} & 32.3 & \textbf{30.1} & \textbf{23.3} & \underline{\textbf{21.8}} \\ \hline  \addlinespace[1pt] 
  & \multicolumn{10}{c}{\textbf{\textit{OTR Score \% $\downarrow$}}}      &  \\ 
\textbf{LLaMA}     & 99.5 & 82.5 & 98.6 & 39.1 & 96.7 & 69.1 & 97.6 & 59.6 & 89.0 & 98.1 & \underline{83.0} \\
\textbf{LLaMA-MT}  & 22.4 & ~~0.5  & 96.9 & ~~3.5  & 18.3 & ~~0.4  & 83.5 & ~~0.0  & 93.8 & 99.5 & \underline{41.9} \\
\textbf{Post-Ins}  & 11.0 & ~~0.7  & 86.8 & ~~3.8  & ~~4.6  & ~~0.5  & 71.7 & ~~0.0  & 68.5 & 99.4 & \underline{34.7} \\
\textbf{PTL}       & 31.8 & ~~0.5 & 98.5 & ~~3.5 & 32.1 & ~~0.4 & \textbf{12.9} & ~~0.0 & 84.3 & 74.4 & \underline{33.8} \\
\textbf{1-shot}    & 24.4 & ~~0.8 & 98.9 & ~~4.4 & 22.4 & ~~0.7 & 50.5 & ~~0.0 & 98.1 & 99.6 & \underline{40.0} \\
\textbf{5-shot}    & 22.2 & ~~0.6 & 96.8 & ~~3.7 & 20.8 & ~~0.6 & 33.1 & ~~0.1 & 99.0 & 99.8 & \underline{37.7} \\
$\mathcal{C}_{lang}$  & ~~5.9 & ~~\textbf{0.4} & \textbf{83.9} & ~~\textbf{2.6} & ~~4.5 & ~~0.4 & 38.1 & ~~0.0 & 86.4 & 91.7 & \underline{31.4} \\ 
\rowcolor{green!10} \textbf{Ours}      & ~~\textbf{5.1}  & ~~0.8  & 88.6 & ~~3.5  & ~~\textbf{2.7}  & ~~\textbf{0.4}  & 15.4 & ~~\textbf{0.0}  & ~~\textbf{1.2}  & ~~\textbf{2.2}  & \underline{\textbf{12.0}} \\ \hline  \addlinespace[1pt] 
  & \multicolumn{10}{c}{\textbf{\textit{BLEURT Score \% $\uparrow$}}}     &  \\
\textbf{LLaMA}     & 26.0  & 46.3 & \textbf{34.6} & 29.5 & 23.0 & 31.8 & 13.3 & 27.2 & 34.9 & 38.2 & \underline{30.5} \\
\textbf{LLaMA-MT}  & 58.4 & 71.1 & 27.3 & 54.5 & 57.8 & 72.3 & 35.1 & 70.8 & 28.9 & 21.5 & \underline{49.8} \\
\textbf{Post-Ins}  & 65.2 & \textbf{71.3} & 28.8 & \textbf{55.5} & \textbf{66.6} & \textbf{73.0} & 43.0 & 71.2 & 40.6 & 21.9 & \underline{53.7} \\
\textbf{PTL}       & 53.2 & 71.1 & 25.7 & 54.5 & 49.3 & 72.3 & 59.3 & 70.8 & 33.0 & 31.4 & \underline{52.1} \\
\textbf{1-shot}    & 56.4 & 70.0 & 26.7 & 52.2 & 55.2 & 71.3 & 44.4 & 69.2 & 27.9 & 21.2 & \underline{49.5} \\
\textbf{5-shot}    & 58.6 & 69.5 & 27.0 & 51.0 & 56.0 & 70.9 & 50.2 & 69.1 & 27.5 & 21.1 & \underline{50.1}  \\
$\mathcal{C}_{lang}$  & 35.5 & 69.1 & 19.9 & 53.0 & 39.5 & 70.7 & 33.1 & 69.3 & 22.4 & 17.2 & \underline{43.0} \\ 
\rowcolor{green!10} \textbf{Ours}      & \textbf{67.1} & \textbf{71.3} & 28.9 & 55.4 & 66.5 & 72.9 & \textbf{61.2} & \textbf{71.3} & \textbf{68.5} & \textbf{59.2} & \underline{\textbf{62.2}} \\ 
\bottomrule
\end{tabular}}
\caption{\textbf{Zero-shot translation performance achieved on WMT benchmark.} \textbf{Bold}: the better results. \underline{Underline}: average scores obtained for all directions.}
\end{table*}

\begin{table*}[t]
\centering
\label{tab:iwslt}
\renewcommand{\arraystretch}{1.15} 
\resizebox{0.9\linewidth}{!}{
\begin{tabular}{lccccccc}
\toprule
\textbf{Models} & \bf Zh$\leftarrow$De & \bf Zh$\rightarrow$De    & \bf Zh$\leftarrow$Ko & \bf Zh$\rightarrow$Ko    & \bf De$\leftarrow$Ko & \bf De$\rightarrow$Ko    & \textbf{AVG}  \\ \hline  \addlinespace[1pt] 
   & \multicolumn{6}{c}{\textbf{\textit{SacreBLEU Score $\uparrow$}}}&   \\ 
\textbf{LLaMA}    & ~~0.1& ~~0.4& ~~0.2& ~~0.2& ~~0.4& ~~0.3& ~~\underline{0.3}      \\ 
\textbf{LLaMA-MT} & ~~0.9& ~~1.0& ~~0.1& ~~0.6& ~~0.7& ~~2.4& ~~\underline{0.9}      \\
\textbf{Post-Ins} & ~~9.1  & ~~6.2  & ~~0.2  & ~~3.1  & ~~3.2  & ~~\textbf{4.1} & ~~\underline{4.3}  \\
\textbf{PTL}      & ~~9.9 & ~~1.4 & ~~0.6 & ~~0.5 & ~~0.7 & ~~1.0 & ~~\underline{2.3} \\
\textbf{1-shot}   & ~~2.5 & ~~1.2 & ~~0.1 & ~~0.9 & ~~0.7 & ~~3.4 & ~~\underline{1.5} \\
\textbf{5-shot}   & ~~6.0 & ~~2.1 & ~~0.2 & ~~1.4 & ~~0.9 & ~~5.0 & ~~\underline{2.6}  \\
$\mathcal{C}_{lang}$ &  ~~0.8 & ~~0.2 & ~~0.1 & ~~0.1 & ~~0.1 & ~~0.4 & ~~\underline{0.3} \\
\rowcolor{green!10} \textbf{Ours}     & \textbf{11.5}   & ~~\textbf{7.1} & ~~\textbf{6.4 }& ~~\textbf{3.5} & ~~\textbf{4.9} & ~~3.3 & ~~\underline{\textbf{6.1 }}     \\\hline  \addlinespace[1pt] 
& \multicolumn{6}{c}{\textbf{\textit{OTR Score \% $\downarrow$}}}    &  \\
\textbf{LLaMA}    & 100.0~~ & 99.8 & 100.0~~ & 99.7 & 99.4 & 99.7 & \underline{99.8}     \\
\textbf{LLaMA-MT}     & 92.9 & 99.5 & 100.0~~ & 99.2 & 99.9 & 63.7 & \underline{92.5}     \\
\textbf{Post-Ins} & 28.6 & 31.9 & 99.7 & 46.9 & 63.6 & ~~6.1 & \underline{46.1} \\
\textbf{PTL}      & 16.6 & 94.0 & 94.3 & 99.8 & 99.4 & 90.7 & \underline{82.5} \\
\textbf{1-shot}   & 13.3 & 23.0 & 13.3 & 15.6 & 20.3 & 26.9 & \underline{20.5} \\
\textbf{5-shot}   & 33.4 & 29.2 & 14.0 & 18.4 & 22.1 & 31.8 & \underline{24.8} \\
$\mathcal{C}_{lang}$ & 82.9 & 94.9 & 99.1 & 97.6 & 97.4 & 44.6 & \underline{86.1} \\
\rowcolor{green!10} \textbf{Ours}     & ~~\textbf{0.8} & ~~\textbf{0.0} & ~~\textbf{0.6} & ~~\textbf{0.0} & ~~\textbf{0.3} & ~~\textbf{0.0} & ~~\underline{\textbf{0.3}}      \\ \hline \addlinespace[1pt] 
& \multicolumn{6}{c}{\textbf{\textit{BLEURT Score \% $\uparrow$}}}&\\ 
\textbf{LLaMA}    & 42.8 & 31.8 & \textbf{46.1} & \textbf{34.3} & 36.6 & \textbf{34.5} & \underline{37.7}     \\
\textbf{LLaMA-MT}     & 20.0 & 22.5 & 13.2 & 14.2 & 19.8 & 23.1 & \underline{18.8}     \\
\textbf{Post-Ins} & 40.0 & 48.4 & 14.2 & 25.9 & 32.1 & \textbf{34.5} & \underline{32.5} \\
\textbf{PTL}      & 38.8 & 24.6 & 15.4 & 14.7 & 20.0 & 16.8 & \underline{21.7} \\
\textbf{1-shot}   & 14.0 & 23.0 & 13.3 & 15.6 & 20.3 & 26.9 & \underline{20.5} \\
\textbf{5-shot}   & 33.4 & 29.2 & 14.0 & 18.4 & 22.1 & 31.8 & \underline{24.8} \\
$\mathcal{C}_{lang}$ & 15.8 & 17.9 & 14.7 & 11.6 & 16.3 & 10.6 & \underline{14.5} \\ 
\rowcolor{green!10} \textbf{Ours}     & \textbf{46.1} & \textbf{57.2} & 34.4 & 31.7 & \textbf{48.4} & 32.9 & \underline{\textbf{41.8}}    \\
\bottomrule
\end{tabular}}
\caption{\textbf{Zero-Shot translation performance on the IWSLT dataset.} \textbf{Bold}: the best results. \underline{Underline}: average scores obtained for all directions.}
\end{table*}

\begin{table*}[t]
\centering
\renewcommand{\arraystretch}{1.2} 
\scalebox{0.9}{
\begin{tabular}{ccccccccc}
\toprule
\textbf{Models} & \textbf{Size} & \bf Zh$\leftarrow$De & \bf Zh$\rightarrow$De    & \bf Zh$\leftarrow$Ko & \bf Zh$\rightarrow$Ko    & \bf De$\leftarrow$Ko & \bf De$\rightarrow$Ko    & \textbf{AVG}  \\ \hline \addlinespace[1pt] 
& & \multicolumn{6}{c}{\textit{\textbf{OTR Score \% $\downarrow$}}}  & \\
\multirow{2}{*}{\textbf{LLaMA-MT}} & \textbf{7B} & 92.9 & 99.5 & 100.0~~ & 99.2 & 99.9 & 63.7& \underline{92.5}  \\
& \textbf{13B} & 30.8 & 68.6 & 99.9  & 99.8 & 81.0 & 40.2& \underline{70.1}  \\  
\rowcolor{green!10} & \textbf{7B} & \bf~~0.8  & \bf~~0.0  & \bf~~0.6   & \bf~~0.0  & \bf~~0.3  & \bf~~0.0 & ~~\underline{\textbf{0.3}}  \\
\rowcolor{green!10} \multirow{-2}{*}{\textbf{Ours}}  & \textbf{13B} & \bf~~1.7 & \bf~~0.0 & \bf~~2.4 & \bf~~0.0 & \bf~~0.3 & \bf~~0.1 & \bf~~\underline{\textbf{0.7}} \\ \hline \addlinespace[1pt] 
& & \multicolumn{6}{c}{\textit{\textbf{BLEURT Score \% $\uparrow$}}}  & \\
\multirow{2}{*}{\textbf{LLaMA-MT}}   & \textbf{7B}  & 20.1 & 22.5  & 13.2 & 14.2  & 19.8 & 23.1  & \underline{18.8} \\
 & \textbf{13B} & 42.6 & 36.1  & 14.4 & 14.6 & 28.1 & 31.9  & \underline{27.9} \\ 
\rowcolor{green!10} & \textbf{7B}  & \bf46.1 & \bf57.2  & \bf34.4 & \bf31.7  & \bf48.4 & \bf33.0  & \underline{\textbf{41.8}} \\
\rowcolor{green!10} \multirow{-2}{*}{ \textbf{Ours}} & \textbf{13B} & \bf49.3 & \bf62.2 & \bf38.7 & \bf31.6 & \bf52.6 & \bf 37.8 &\underline{\textbf{45.4}} \\
\bottomrule
\end{tabular}}
\caption{\textbf{The impact of model size.} We report the BLEURT and OTR Scores on the IWSLT dataset. \textbf{Bold}: the better results of LLaMA-MT and ours. \underline{Underline}: average scores obtained for all directions.}
\label{tab:ablation_over_model_size}
\end{table*}

\subsection{Main Results}
We present the zero-shot translation performance comparison of our model and other baselines on WMT and IWSLT datasets, as depicted in Table~\ref{tab:iwslt} and Table~\ref{tab:wmt}. Our model outperforms the considered baselines across 16 translation directions.

Compared with LLaMA-MT, which is only tuned on multilingual translation data, our model significantly reduces the average OTR scores by -29.9\% in WMT and -92.9\% in IWSLT through unlikelihood training on instruction-conflicting samples. 
In contrast to the baseline approaches that focus on mitigating off-target problems during inference, such as PTL, $K$-shot, and $\mathcal{C}_{lang}$, 
our model demonstrates superior performance, achieving improvements up to +11.3/ +6.2 SacreBLEU, -28.0\%/ -40.2\% OTR, and +19.3/ +21.3 BLEURT in IWSLT/ WMT datasets. 
Regarding baseline adjustments during the tuning stage, our model achieves significant improvements over Post-Ins, average +1.8/ +4.0 SacreBLEU score, -45.8\%/ -22.7\% OTR score, and +9.3/ +8.5 BLEURT score in IWSLT/ WMT datasets. 
Additionally, our model surpasses other robust baseline models in these evaluations.



\section{Analysis}

\subsection{Effect of Unlikelihood Training Steps}

To provide insight into the impact of unlikelihood training steps, Figure~\ref{fig:Ablation}. a) presents the zero-shot translation performance on the IWSLT dataset. 
As observed, the model produces fewer wrong language translations and higher quality translations with more unlikelihood training steps. 
From the figure, it can seen that The model achieves the best performance, denoted by near-zero OTR scores, after about 60 updates, and this performance is consistently maintained even with further training extending up to 100 steps. 

\subsection{Effect of \texorpdfstring{$\alpha$}{Lg}}

As mentioned in Section~\ref{sec:ul}, our algorithm has a mixing hyper-parameter $\alpha$ to balance MLE loss and UL loss. 
This is an ablation to evaluate the effect of different $\alpha$. 

\begin{figure*}[!htb] 
\centering
\subfigure{
\begin{minipage}[b]{0.45\linewidth}%
\setlength{\abovecaptionskip}{0pt}
\begin{center}
     \includegraphics[width=1.0\linewidth]{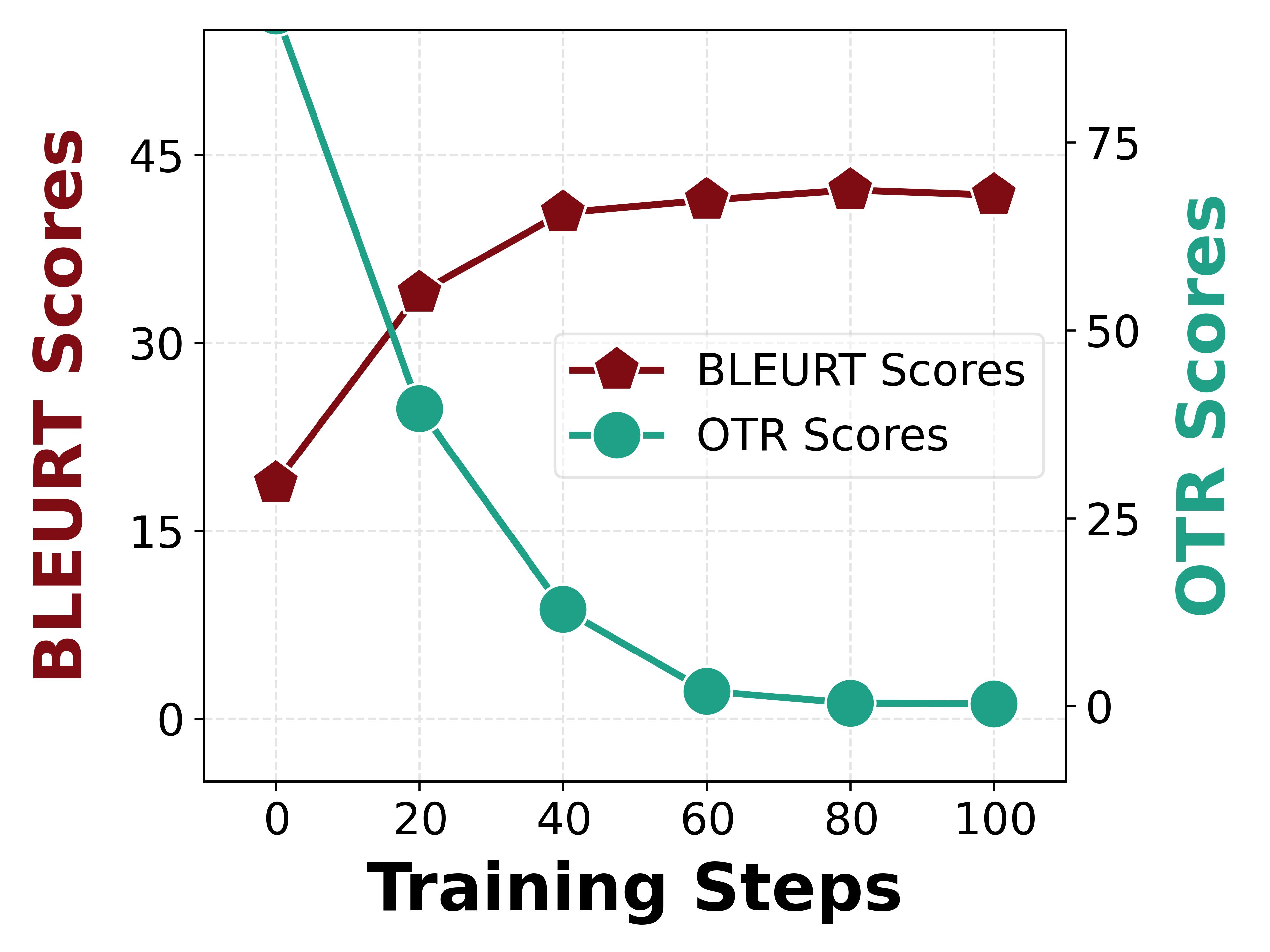}
     \centerline{a) Ablation over steps}
\end{center}
\end{minipage}
}
\subfigure{
\begin{minipage}[b]{0.45\linewidth}%
\setlength{\abovecaptionskip}{0pt}
\begin{center}
     \includegraphics[width=0.8\linewidth]{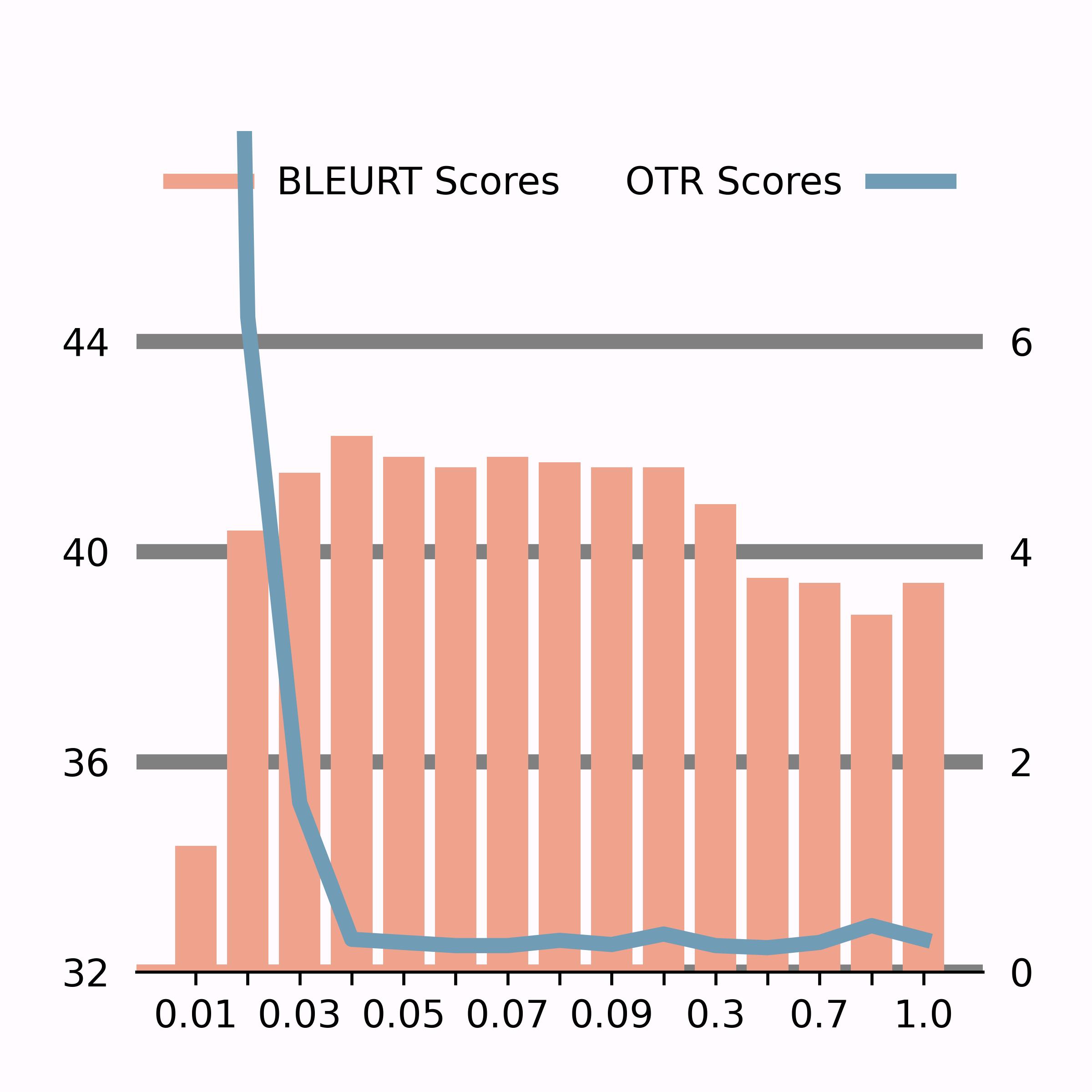}
     \centerline{b) Ablation over $alpha$}
\end{center}
\end{minipage}
}
\caption{\textbf{Ablation Studies.} a) Ablation study on continued training steps. b) Ablation study on the mixing hyper-parameter $\alpha$. This demonstrates the zero-shot translation performance following the second stage of training.}
\label{fig:Ablation}
\end{figure*}

Figure~\ref{fig:Ablation}. b) shows the performance on the IWSLT dataset.
As expected, the higher $\alpha$ highlights the UL loss, resulting in fewer wrong language translations. Models fine-tuned with $\alpha$ exceeding 0.04 are unlikely to produce translations in wrong language.
However, when $\alpha$ is increased beyond 0.3, there is a slight decrease in translation quality (with BLEURT scores of 42.2 vs. 38.8). This decline may be attributed to potential overfitting on the unlikelihood loss. Future research efforts should be directed toward mitigating the effects of this potential overfitting issue.
In summary, our experimental results indicate that our method exhibits robustness to varying values of the mixing parameter, $\alpha$.

\subsection{Results with Different Size of LLMs}
To investigate the influence of model size, we conducted experiments with 13B size LLaMA on the multilingual translation dataset, employing the same experimental setup as that of the 7B size model
The results are summarized in Table~\ref{tab:ablation_over_model_size}. The 13B model consistently outperforms the 7B model in terms of both the reduction in wrong language translations (-22.4\% average OTR) and the improvement in translation quality (+9.1 average BLEURT). This observation aligns with prior findings~\cite{kaplan2020scaling}, which suggest that increasing the number of training parameters yields benefits.
However, the off-target problem still exists in the 13B size LLaMA-MT model. 
Our model achieves a significantly lower off-target translation ratio(0.7\% vs. 70.1\% average OTR), leading to a higher quality translation (45.4 vs. 27.9 average BLEURT). 
This result demonstrates that our algorithm remains effective with larger LLMs.

\subsection{Results with Different Amounts of Translation Data}
\begin{wrapfigure}{r}{0.5\textwidth}
\vspace{-20pt}
\centering
\includegraphics[width=0.46\textwidth]{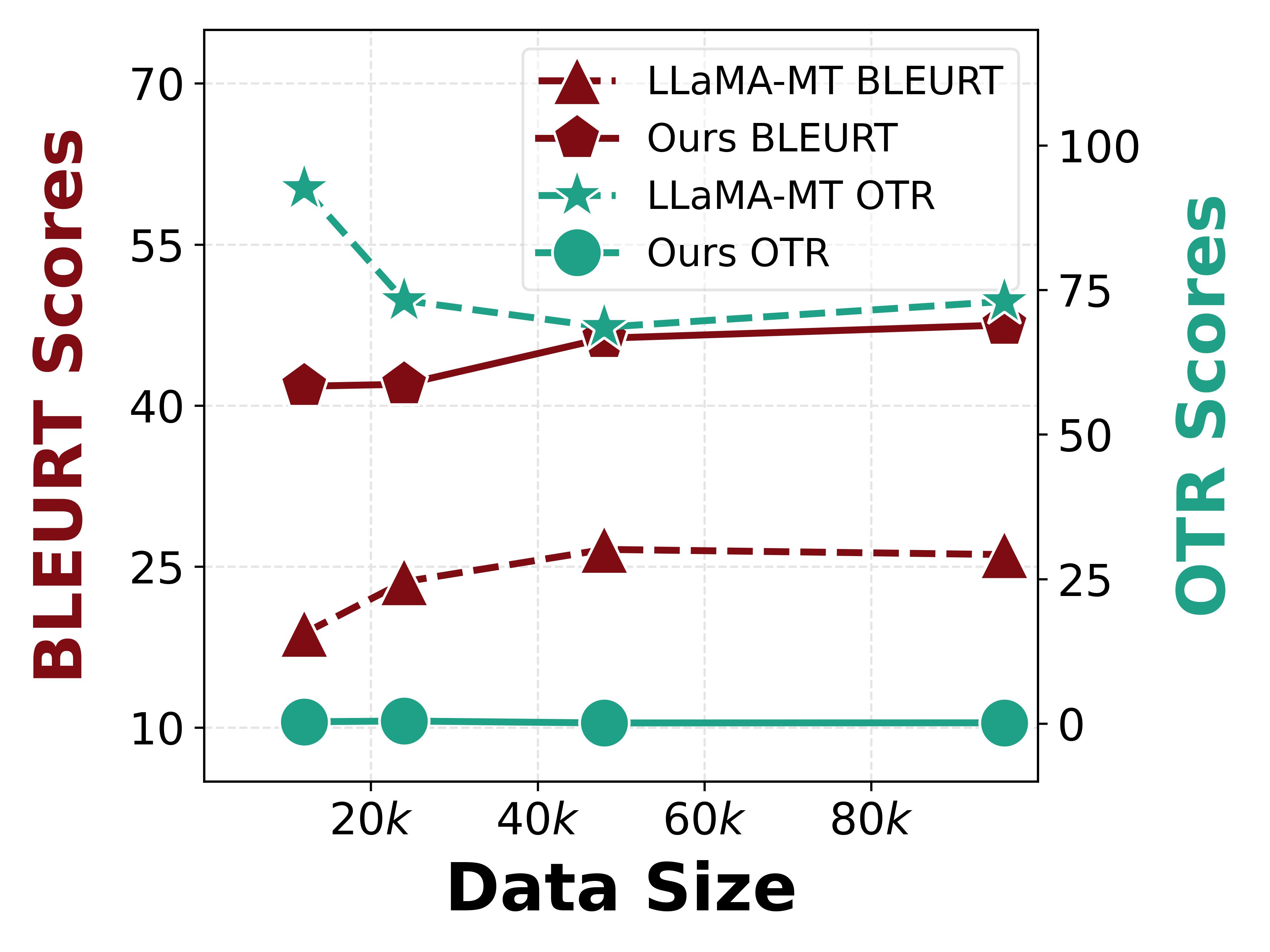}
    \caption{
    \textbf{The impact of fine-tuning translation data size.} We report the BLEURT and OTR Scores on the IWSLT dataset. The x-axis is the fine-tuning data size $n$.}
    \label{fig:ablation_over_fine-tuning_data_size}
\end{wrapfigure}

Figure~\ref{fig:ablation_over_fine-tuning_data_size} illustrates the zero-shot translation performance of our models, according to  BLEURT and OTR score, with a pre-tuning multilingual translation dataset consisting of $n$ samples.
Although we consider the zero-shot translation performance, it also brings gains with more translation data. 
However, the benefits derived from augmenting the translation data size become negligible when the dataset exceeds 40k samples.
Additionally, our algorithm exhibits robustness to different translation data sizes and consistently achieves OTR scores close to zero across all four settings, consequently leading to significantly higher BLEURT scores.

\subsection{Performance on Supervised Translation}
\begin{table*}[t]
\centering
\renewcommand{\arraystretch}{1.2} 
\scalebox{0.95}{
\begin{tabular}{lcccccc}
\toprule
\multirow{2}{*}{\textbf{Models}} & \multicolumn{3}{c}{\textbf{SacreBLEU$\uparrow$}}  & \multicolumn{3}{c}{\textbf{BLEURT$\uparrow$}} \\ \cmidrule(lr){2-4} \cmidrule(lr){5-7}
& \textbf{IWSLT} & \textbf{WMT} & \textbf{AVG}  & \textbf{IWSLT-4} & \textbf{WMT-3} & \textbf{AVG}   \\ \hline \addlinespace[1pt] 
\textbf{LLaMA-MT} & \textbf{16.7}    & 23.4 & \textbf{\underline{20.0}} & \textbf{57.5}   & 65.4 & \underline{61.4}\\
\rowcolor{green!10} \bf Ours & 15.9   & \textbf{23.5}  & \underline{19.7} & 57.4  & \textbf{65.7}  & \textbf{\underline{61.6}} \\
\bottomrule
\end{tabular}}
\caption{\textbf{Supervised translation performance.} \textbf{Bold}: the best results. \underline{Underline}: average scores.}
\label{tab:supervised_performance}
\end{table*}

\begin{table*}[t]
\centering
\renewcommand{\arraystretch}{1.15} 
\scalebox{0.95}{
\begin{tabular}{lccccc}
\hline
& \bf Win Rate \% $\uparrow$ & \bf SacreBLEU $\uparrow$ & \bf OTR \% $\downarrow$ & \bf BLEURT $\uparrow$ \\ \hline
\textbf{Alpaca-MT} & \textbf{45.4} & ~~3.7  & 47.1  & 28.7   \\
\rowcolor{green!10}\textbf{Ours}       & 43.3 & ~~\textbf{5.0}  & ~~\textbf{4.8}  & \textbf{29.5}  \\
\hline
\end{tabular}}
\caption{\textbf{Performance after combining with general tasks data.} We report the Win rate \% compared with Alpaca and translation performance on IWSLT. \textbf{Bold}: the best results.}
\label{tab:general_performance}
\end{table*}
As our algorithm primarily enhances zero-shot translation performance through the unlikelihood training on instruction-conflicting samples. This raises a question: does the supervised translation ability persist even after unlikelihood training?
As shown in Table~\ref{tab:supervised_performance}, we report the performance of LLaMA-MT and ours on IWSLT and WMT. 
Remarkably, our models successfully retain the supervised translation ability after unlikelihood training with instruction-conflicting samples. 
Specifically, our final model achieves comparable results compared with LLaMA-MT (19.7 vs 20.0 in SacreBLEU score and 61.6 vs. 61.4 in BLEURT score).

\subsection{Effect on General Task Performance}
Inspired by \citet{jiao2023parrot}, we consider improving the zero-shot translation capabilities of LLMs tuned on a mixed dataset, consisting of translation data and general task data. 
We construct the instruction tuning dataset by combining Alpaca\footnote{\url{https://github.com/tatsu-lab/stanford_alpaca}} with IWSLT translation samples and using the same hyperparameters as the main experiments for training. 
Following AlpacaEval\footnote{\url{https://github.com/tatsu-lab/alpaca_eval}}, we assess the performance with GPT-4~\cite{openai2023gpt4} as the evaluator, while taking the reproduced Alpaca as the reference model to compute the win rate \%. 
As shown in Table~\ref{tab:general_performance}, our model attains comparable general tasks performance with Alpaca-MT (43.3\% vs. 45.4\%), while boosting the zero-shot translation performance (+1.3 SacreBLEU, -42.3\% OTR, and +0.8 BLEURT), confirming the effectiveness of our algorithm when employed with a general task dataset.

\section{Related Work} 
\label{related_work}

\paragraph{Translation-Tailored LLMs}
Due to the huge cost to call the state-of-the-art LLMs, such as GPT-4~\cite{openai2023gpt4}, there is a need to investigate how to effectively fit a smaller LLM into specific tasks, e.g., machine translation. Note that although there are some powerful sequence-to-sequence style large-scale pretrained machine translation models~\cite{Liu2020MultilingualDP,zan2022vega}, this paper mainly focuses on the decoder-only LLMs due to their flexible interaction modes and rich world knowledge.
In the field of LLMs-based translation, various approaches have been proposed to optimize translation performance. 
Parrot~\cite{jiao2023parrot} proposes to fine-tune model on machine translation data with a hint incorporating extra requirements to regulate the translation process. 
TIM~\cite{zeng2023tim} introduces translation samples in comparisons to compute additional preference loss for regularization, exhibiting superior translation ability in both supervised and zero-shot directions. 
ALMA~\cite{xu2023paradigm_shift} proposes a two-stage approach that first fine-tunes on monolingual data of downstream languages followed by fine-tuning on high-quality translation data, which achieves significant improvement of translation quality. 
\citet{liu2023instruction_position} presents the position of instruction matters, that just moving the location of the instruction closer to the output can alleviate the instruction forgetting issue. 

In contrast, we focus on the off-target problem of zero-shot translation, where the model fails to follow translation instructions, generating sequences not in the target language.
Additionally, we show how instruction-conflicting samples can enhance the influence of instruction thus mitigating the off-target problem.

\paragraph{Unlikelihood Training} 
Unlikelihood training~\cite{welleck2019unlikelihood_training} aims to force the mode to assign a lower probability for unlikely tokens. 
This method has been further explored in dialog tasks by \citet{li-etal-2020-dont}, who demonstrated its effectiveness in generating more consistent and coherent human-like dialog.
\citet{nogueira-dos-santos-etal-2020-beyond} used the unlikelihood loss for ranking and proposed a generative information retrieval approach.
\citet{hosseini-etal-2021-understanding} proposed the combination of an unlikelihood objective with a reference-based setup for input sentences to model negation with pretrained BERT~\cite{kenton2019bert}.
\citet{hu-etal-2023-uncertainty-ul} take the semantic-similar or ambiguous
tokens as negative information and acquire it via inherent uncertainty for the ASQP task. 

In this work, we take instances in which the translation pairs conflict with the instruction as the negative sample for zero-shot translation. 
Furthermore, we consider the new case that enhances the ability of LLMs to better follow translation instructions and generate translations in the correct language. 
\section{Conclusion}
We propose a simple two-stage finetuning strategy to enhance the instruction-following ability of LLM for translation. 
The core procedure consists of two main steps: 1) creating instruction-conflicting samples by replacing the translation directions with incorrect ones, and 2) training on these samples using an additional unlikelihood loss.
Experimental results on IWSLT and WMT, spanning 16 zero-shot translation directions, demonstrate the effectiveness of the proposed method, which reduces the off-target translation ratio and produces translations with higher quality.
Furthermore, our approach exerts a negligible influence on other aspects of LLMs, such as supervised translation performance and general task performance.

\section*{Limitations}
The proposed method contains a mixing hyperparameter $\alpha$ to balance MLE loss and UL loss in unlikelihood training on instruction-conflicting samples, And, the high $\alpha$ may overfit the model on UL loss. In future work, we may focus on how to balance them in an adaptive way. 

This work only focuses on the off-target problem in the zero-shot translation of LLMs, which could be seen as a specific type of input-conflicting hallucination. 
In future work, we will continue to explore the application of the unlikelihood training on general tasks, such as programming, math, dialogue, and etc, and more types of hallucinations~\cite{zhang2023siren}, such as fact-conflicting hallucinations, and context-conflicting hallucinations. Also, we will apply our method to enhance the instruction following abilities of different interesting LLMs-based scenarios, e.g., safety~\cite{zhang2024intention,zhong2024rose}, debiasing~\cite{xu2024promptbias}, multimodal analysis~\cite{wang2024wisdom}, healthcare~\cite{ren2024healthcare}, and difficult code generation~\cite{wang2024oop}.

\section*{Ethical Consideration}
We take ethical considerations very seriously and strictly adhere to the ethics policy. 
In this paper, we aim to improve the zero-shot translation ability in LLMs fine-tuned with translation using publicly available and widely-used datasets. 
The results and conclusions presented are done with accuracy and objectivity.

\bibliographystyle{plainnat}
\bibliography{main}

\end{document}